\pdfoutput=1

\documentclass[11pt]{article}

\usepackage[]{EMNLP2023}

\usepackage{times}
\usepackage{latexsym}

\usepackage[T1]{fontenc}

\usepackage[utf8]{inputenc}

\usepackage{microtype}

\usepackage{inconsolata}
\usepackage{times}
\usepackage{latexsym}
\usepackage{CJKutf8}
\usepackage[tone]{tipa}
\usepackage{tipa}
\usepackage{amsmath}
\usepackage{amssymb}
\usepackage{tikz}
\usepackage{ctable}
\usepackage{graphicx}
\usepackage{caption}
\usepackage{balance}
\usepackage{subfig}
\usepackage{arydshln}
\usepackage{balance}
\usepackage{multirow}

\title{CoF-CoT: Enhancing Large Language Models with Coarse-to-Fine Chain-of-Thought Prompting for Multi-domain NLU Tasks}

\author{Hoang H. Nguyen$^{1}$, Ye Liu$^{2}$, Chenwei Zhang$^{3}$, Tao Zhang$^{1}$, Philip S. Yu$^{1}$\\
  $^1$ Department of Computer Science, University of Illinois at Chicago, Chicago, IL, USA \\
  $^2$ Salesforce Research, Palo Alto, CA, USA \\
  $^3$ Amazon, Seattle, WA, USA \\
  \texttt{\{hnguy7,tzhang90,psyu\}@uic.edu, yeliu@salesforce.com, cwzhang@amazon.com}
  }

\begin{document}
\maketitle
\begin{abstract}

While Chain-of-Thought prompting is popular in reasoning tasks, its application to Large Language Models (LLMs) in Natural Language Understanding (NLU) is under-explored. Motivated by multi-step reasoning of LLMs, we propose Coarse-to-Fine Chain-of-Thought (CoF-CoT) approach that breaks down NLU tasks into multiple reasoning steps where LLMs can learn to acquire and leverage essential concepts to solve tasks from different granularities. Moreover, we propose leveraging semantic-based Abstract Meaning Representation (AMR) structured knowledge as an intermediate step to capture the nuances and diverse structures of utterances, and to understand connections between their varying levels of granularity. Our proposed approach is demonstrated effective in assisting the LLMs adapt to the multi-grained NLU tasks under both zero-shot and few-shot multi-domain settings \footnote{\href{https://github.com/nhhoang96/CoF-CoT}{https://github.com/nhhoang96/CoF-CoT}}. 
\end{abstract}
\section{Introduction}
\vspace*{-0.2cm}
Natural Language Understanding (NLU) of Dialogue systems encompasses tasks from different granularities. Specifically, while intent detection requires understanding of coarse-grained sentence-level semantics, slot filling requires fine-grained token-level understanding. 
Moreover, Semantic Parsing entails the comprehension of connections between both token-level and sentence-level tasks. 

Large Language Models (LLMs) possess logical reasoning capability and have yielded exceptional performance \cite{52065, zhao2023survey}. However, they remain mostly restricted to reasoning tasks. On the other hand, mutli-step reasoning can take place when solving multiple interconnected tasks in a sequential order.
In practical NLU systems, as coarse-grained tasks are less challenging, they can be solved first before proceeding to fine-grained tasks. Therefore, the coarse-grained tasks' outcomes can provide valuable guidance towards subsequent fine-grained tasks,
 allowing for deeper semantic understanding of diverse utterances across different domains within NLU systems \cite{firdaus2019multi, weld2022survey, nguyen-etal-2023-slot}. For instance, consider the utterance \textit{``Remind John the meeting time at 8am''} under \textbf{reminder} domain, recognizing \textbf{GET\_REMINDER\_DATE\_TIME} intent is crucial for correctly understanding the existence of \textbf{PERSON\_REMINDED} slot type rather than \textbf{CONTACT} or \textbf{ATTENDEE} slot type.



\begin{figure}[bt]
    \centering
   \includegraphics[trim={4.0cm 6.0cm 5.5cm 1.5cm},clip, width=\columnwidth]{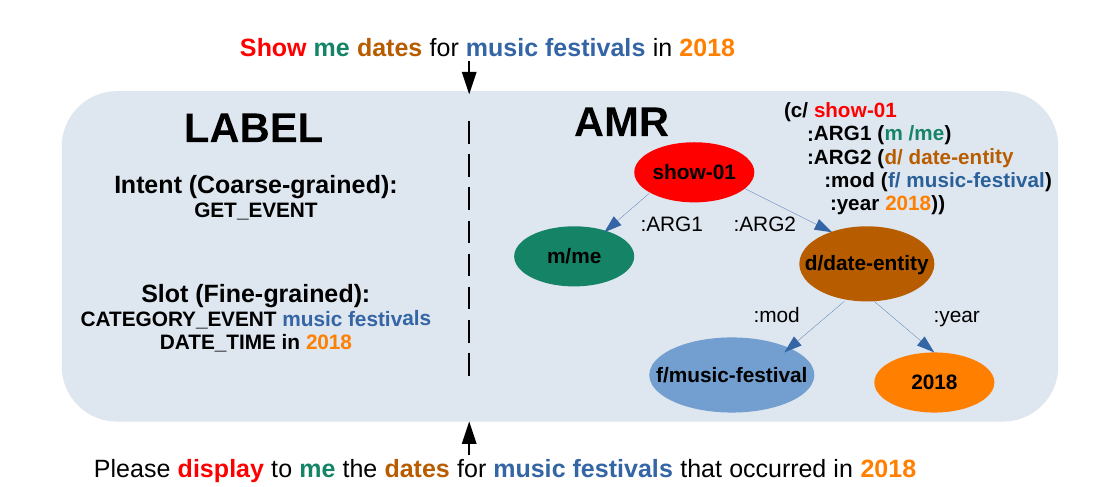}\
   \vspace*{-0.6cm}
    \caption{Illustration of \textbf{Abstract Meaning Representation (AMR)} of two structurally different but semantically similar utterances with the same fine-grained and coarse-grained labels. 
    Each colored node represents an AMR concept matching the colored word or phrase existent in the corresponding utterances. 
    }
    \label{fig:example}
    \vspace*{-0.7cm}
\end{figure} 
\begin{figure*}[t]
    \centering
   \includegraphics[trim={0.4cm 3.7cm 0.5cm 0.5cm},clip, width=0.9\textwidth]{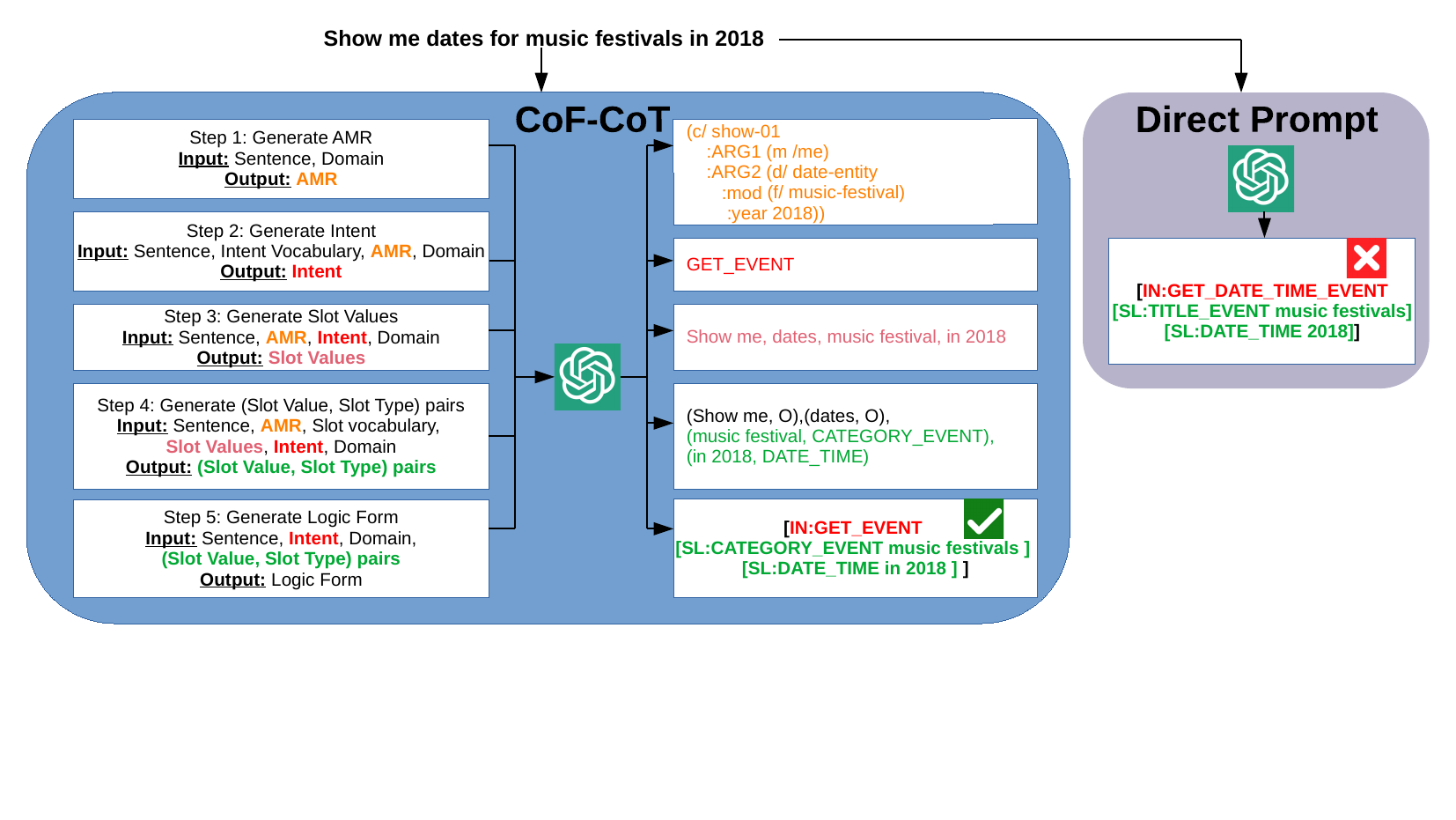}
   \vspace*{-0.2cm}
    \caption{Illustration of \textbf{CoF-CoT} and its counterpart \textbf{Direct Prompt} approach. The left side illustrates the proposed CoF-CoT. The right side illustrates the naive Direct Prompt approach. \textbf{\textcolor{red}{Red}} and \textbf{\textcolor{green}{Green}} represent \textbf{\textcolor{red}{sentence-level}} and \textbf{\textcolor{green}{token-level}} annotations captured in the Logic Form respectively. For CoF-CoT, the prompt at each step starting from Step 2 is conditioned on the relevant output predicted from the previous step(s). 
    }
    \label{fig:overview}
    \vspace*{-0.6cm}
\end{figure*} 

Chain-of-Thought (CoT) \cite{wei2022chain} provides an intuitive approach to elicit multi-step reasoning from LLMs automatically. However, there remain two major challenges with the current CoT approach: (1) LLMs entirely rely on their uncontrollable pre-trained knowledge to generate step-by-step reasoning and could result in unexpected hallucinations \cite{yao2022react,zhao2023verify}, (2) Additional beneficial structured knowledge cannot be injected into LLMs via the current CoT. 

On the other hand, structured representation demonstrates the effectiveness in enhancing the capability of Pre-trained Language Models (PLMs) \cite{xu2021syntax, bai-etal-2021-syntax,shou-etal-2022-amr}. In Dialogue systems, the dependencies among different dialogue elements together with the existent diversely structured utterances necessitate the integration of additional structured representation. For instance, as observed in Figure \ref{fig:example}, by leveraging Abstract Meaning Representation (AMR) \cite{banarescu2013abstract}, it is possible to map multiple semantically similar but structurally different utterances with similar coarse-grained and fine-grained labels into the same structured representation, allowing for effective extraction of intents, slots, and their interconnections within the Dialogue systems.    


In our work, we explore the capability of LLMs in NLU tasks from various granularities, namely multi-grained NLU tasks. Motivated by CoT, we propose an adaptation of CoT in solving multi-grained NLU tasks 
with an integration of structured knowledge from AMR Graph. Our contribution can be summarized as follows:

$\bullet$ To the best of our knowledge, we conduct the first preliminary study of LLMs' capability in multi-grained NLU tasks of the Dialogue systems.\\
\indent $\bullet$ We propose leveraging a CoT-based approach to solve multi-grained NLU tasks in a coarse-to-fine-grained sequential reasoning order. \\
\indent $\bullet$ We propose integrating structured knowledge represented via AMR Graph in the multi-step reasoning to capture the shared semantics across diverse utterances within the Dialogue systems. 
\vspace*{-0.2cm}
\section{Related Work}
\vspace*{-0.2cm}
\paragraph{Chain-of-Thought (CoT)} CoT \cite{wei2022chain} proposes leveraging intermediate steps to extracts logical reasoning of LLMs and succeeds in various reasoning tasks. \citet{wang2022self} enhances CoT by selecting the most consistent output answers via majority voting. Additionally, \citet{fu2022complexity} argues majority consistency voting works best among the most complex outputs. They propose complexity metrics and leverage them to select demonstration samples and decoding outputs. Unlike previous CoT approaches, we leverage CoT to solve multi-grained NLU tasks.
 \vspace*{-0.1cm}
\paragraph{Structured Representation}
Structured Representation has been widely incorporated in language models to further enhance the capability across various NLP tasks \cite{bugliarello-okazaki-2020-enhancing,zhang2020sg}. Structured representation can be either in the syntax-based structure \cite{bai-etal-2021-syntax,xu2021syntax} such as Dependency Parsing (DP) Graph, Constituency Parsing (CP) Graph or semantic-based structure \cite{shou-etal-2022-amr} such as Abstract Meaning Representation (AMR) Graph \cite{banarescu2013abstract}. Unlike previous works, we aim at leveraging structured representation as an intermediate step in the multi-step reasoning approach to extract essential concepts from diverse utterances in the multi-domain Dialogue systems.  
 \begin{table*}[htb]
\centering
\caption{Experimental results on MTOP and MASSIVE under zero-shot and few-shot multi-domain settings.}
\vspace*{-0.3cm}
\resizebox{\textwidth}{!}{%
\begin{tabular}{||c||c|c||c|c||c|c||c|c||}
\hline 
\multicolumn{9}{||c||}{\textbf{MTOP}} \\
\hline 
 \textbf{Model} & \multicolumn{4}{c||}{\textbf{Zero-shot}} & \multicolumn{4}{c||}{\textbf{Few-shot}}\\
\specialrule{.1em}{0.05em}{.05em}
 & \multicolumn{2}{c||}{NLU} & \multicolumn{2}{c||}{Semantic Parsing} 
  & \multicolumn{2}{c||}{NLU} & \multicolumn{2}{c||}{Semantic Parsing} \\
  \hline 
  &  Intent Acc & Slot F1 & Frame Acc & Exact Match & Intent Acc &  Slot F1 & Frame Acc & Exact Match  \\


\specialrule{.1em}{0.05em}{.05em}
Direct Prompt 
 & 31.50 $\pm$ 1.80 & 21.84 $\pm$ 2.83 & 8.33 $\pm$ 1.44 & 6.00 $\pm$ 1.32

 & 51.33 $\pm$ 3.40 & 28.35 $\pm$ 3.24 & 11.00 $\pm$ 1.80 & 8.33 $\pm$ 1.00


 \\

CoT
 & 31.83 $\pm$ 2.02 & 22.40 $\pm$ 1.61 & 8.67 $\pm$ 0.35 & 6.33 $\pm$ 1.04

 & 47.67 $\pm$ 5.20 & 28.46 $\pm$ 3.10 & 11.83 $\pm$ 1.53 & 8.50 $\pm$ 1.04

\\

SC-CoT

& 32.50 $\pm$ 1.89 & 22.71 $\pm$ 2.44 & 10.05 $\pm$ 0.87 & 6.83 $\pm$ 0.76

 & 53.50 $\pm$ 3.04 & 29.53 $\pm$ 1.99 & 12.50 $\pm$ 1.80 & 9.00 $\pm$ 0.87

\\

ComplexCoT

  &  32.67 $\pm$ 2.00 & 22.86 $\pm$ 3.17 & 10.83 $\pm$ 0.29 & 7.16 $\pm$ 0.58

 & 48.83 $\pm$ 2.47 & 29.21 $\pm$ 2.65 & 13.17 $\pm$ 0.58 & 8.83 $\pm$ 2.89
\\

Least-to-Most

  &  45.67 $\pm$ 0.58 & 21.84 $\pm$ 1.91 & \textbf{14.50 $\pm$ 0.50} & 8.00 $\pm$ 0.50

 & 49.83 $\pm$ 4.54 & 27.28 $\pm$ 2.41 & 16.00 $\pm$ 0.50 & 8.83 $\pm$ 0.76
\\

Plan-and-Solve

  &  45.00 $\pm$ 4.00 & 22.45 $\pm$ 2.28 & 9.50 $\pm$ 1.61 & 8.25 $\pm$ 2.25
 & -- & -- & -- & --
\\

\textbf{CoF-CoT}
& \textbf{57.67 $\pm$ 2.75} & \textbf{23.47 $\pm$ 4.09} & 14.33 $\pm$ 1.52 & \textbf{9.00 $\pm$ 1.00} 

 & \textbf{61.50 $\pm$ 4.93} & \textbf{30.12 $\pm$ 3.93} & \textbf{15.00 $\pm$ 1.32} & \textbf{11.00 $\pm$ 1.61} 
\\
\hline
\end{tabular}%
}

\resizebox{\textwidth}{!}{%
\begin{tabular}{||c||c|c||c|c||c|c||c|c||}
\specialrule{.1em}{0.05em}{.05em}

\multicolumn{9}{||c||}{\textbf{MASSIVE}} \\

\specialrule{.1em}{0.05em}{.05em}
Direct Prompt 
& 72.50 $\pm$ 4.58 & 33.24 $\pm$ 3.34 & 24.17 $\pm$ 3.79 & 20.67 $\pm$ 3.28

& 75.17 $\pm$ 0.58 & 42.36 $\pm$ 2.98 & 29.00 $\pm$5.39 & 24.50 $\pm$ 4.07
 \\
 
CoT
& 71.83 $\pm$ 2.57 & 36.32 $\pm$ 1.94 &24.50 $\pm$ 2.29 & 21.66 $\pm$ 3.40

&  76.83 $\pm$ 3.82 & 44.89 $\pm$ 2.50  & 31.33 $\pm$ 0.87 &25.83 $\pm$ 2.25

\\

SC-CoT
 & 73.05 $\pm$ 1.27 & 37.06 $\pm$ 2.54 & 27.16 $\pm$3.21 & 22.50 $\pm$ 2.65

 & 77.33 $\pm$ 2.89 & 47.02 $\pm$ 4.60 & 34.00 $\pm$ 3.21 & 27.16 $\pm$ 3.50

\\

ComplexCoT
  &  73.66 $\pm$ 3.65 & 37.64 $\pm$ 3.51 & 25.83 $\pm$ 2.25 & 22.16 $\pm$ 2.51

& 77.83 $\pm$  1.83 & 46.59 $\pm$ 2.43  & 36.50 $\pm$ 2.89 & 28.00 $\pm$ 3.69
\\

Least-to-Most



 & 72..83 $\pm$ 4.65 & 37.62 $\pm$ 1.69 & 31.50 $\pm$ 1.53 & 26.50 $\pm$ 1.26
 & 77..00 $\pm$ 3.28 & 45.93 $\pm$ 3.99 & 32.50 $\pm$ 4.09 & 29.00 $\pm$ 5.11
\\

Plan-and-Solve

 & 69.33 $\pm$ 2.47 & 38.07 $\pm$ 2.07  & 32.00 $\pm$ 1.26 & \textbf{29.00 $\pm$ 1.26}
 & -- & -- & -- & --
\\

\textbf{CoF-CoT}
& \textbf{89.00 $\pm$ 2.29} & \textbf{38.66 $\pm$ 3.25} & \textbf{33.17 $\pm$ 4.04} & 25.50 $\pm$ 2.64 
 & \textbf{92.00 $\pm$2.29}  & \textbf{47.06 $\pm$ 4.63} & \textbf{37.50 $\pm$ 1.89} & \textbf{29.50 $\pm$ 3.12}
\\
\hline
\end{tabular}%
}
\vspace*{-0.5cm}

\label{tab:main_result}
\end{table*}
 \vspace*{-0.1cm}
\section{Proposed Framework}
\vspace*{-0.2cm}
In this section, we introduce our proposed Coarse-to-Fine Chain-of-thought (CoF-CoT) approach for NLU tasks as depicted in Figure \ref{fig:overview}. Specifically, we propose a breakdown of multi-grained NLU tasks into 5 sequential steps from coarse-grained to fine-grained tasks. At each step, LLMs leverage the information from the previous steps as a guidance towards the current predictions. As domain name could provide guidance to NLU tasks \cite{xie-etal-2022-unifiedskg, zhou-etal-2023-xqa}, at each step, we condition the \textit{domain name} of the given utterance in the input prompt.   The model's output is in the format of Logic Form \cite{kamathsurvey} which encapsulates coarse-grained intent label, fine-grained slot labels and slot values. 
Further details of Logic Form's structure and its connections with multi-grained NLU tasks are provided in the Appendix \ref{ap:logic_form}.

Our multi-step reasoning is designed in the following sequential order:

    \indent 1. \textbf{Generate AMR}: Given the input utterance, LLMs generate the AMR structured representation \cite{banarescu2013abstract}. The representation is preserved in the Neo-Davidsonian format as demonstrated in Figure \ref{fig:example},\ref{fig:overview}. Each node in AMR graph refers to a concept, including entity, noun phrase, pre-defined frameset or special keyword. Edges connecting two nodes represent the relation types.      
    
    \indent 2. \textbf{Generate Intent}: In this step, LLMs generate coarse-grained intent label prediction when conditioned on the given input and its corresponding AMR Graph. AMR concepts could provide additional contexts to ambiguous utterances, leading to improved ability to recognize the correct intents. 
    
    \indent 3.  \textbf{Generate Slot Values}: In this stage, to generate the fine-grained slot values existent in the input utterance, besides the utterance itself, prompts for LLMs are conditioned on the generated AMR structure and predicted intent label. As AMR graph captures the essential concepts existent in the utterance while abstracting away syntactic idiosyncrasies of the utterance, it can help extract the important concepts mentioned in the utterances. 
    In order to further couple the connections between slots and intents \cite{zhang2019joint, wu-etal-2020-slotrefine}, predicted intents from the Step 2 are also concatenated to construct input prompts for Step 3.     
    
    \indent 4.  \textbf{Generate Slot Value, Slot Type pairs}: After obtaining slot values, LLMs label each identified slot value when given the slot vocabulary. Similar to Step 3, we condition the generated output with the predictions from previous steps, including AMR and intent. Both AMR and intent provide additional contexts for slot type predictions of the given slot values besides the input utterance.  
    
    \indent 5.  \textbf{Generate Logic Form}: The last step involves aggregating the predicted intents together with sequences of slot type and slot value pairs to construct the final Logic Form predictions. 

\vspace*{-0.2cm}
\section{Experiments}
\vspace*{-0.2cm}
\subsection{Datasets \& Preprocessing}
\vspace*{-0.1cm}
We evaluate our proposed framework on two multi-domain NLU datasets, namely MTOP \cite{li-etal-2021-mtop} and MASSIVE \cite{ bastianelli-etal-2020-slurp, fitzgerald2022massive}. As the innate capability of language understanding is best represented via the robustness across different domains, we evaluate the frameworks under low-resource multi-domain settings, including zero-shot and few-shot. Details of both datasets are provided in Appendix \ref{ap:dataset}.

To provide a comprehensive evaluation for coarse-grained, fine-grained NLU tasks, as  well as the interactions between the two, we conduct an extensive study on both NLU and Semantic Parsing metrics, including: Slot F1-score, Intent Accuracy, Frame Accuracy, Exact Match. Intent Accuracy assesses the performance on coarse-grained sentence-level tasks, while Slot F1 metric evaluates the performance on more fine-grained token-level tasks. The computation of Frame Accuracy and Exact Match captures the ability to establish the accurate connections between sentence-level and token-level elements. For more details of individual metric computation from the Logic Form, we refer readers to \cite{li-etal-2021-mtop}. 

To conduct the evaluation with efficient API calls, following \cite{khattab2022demonstrate}, we construct test sets by randomly sampling 200 examples covering a set of selected domains, namely  test domains. We repeat the process with 3 different seeds to generate 3 corresponding test sets. Reported performance is the average across 3 different seed test sets with standard deviations. For few-shot settings, we randomly select a fixed k samples from a disjoint set of domains, namely train domains. These samples are manually annotated with individual step labels as commonly conducted by other in-context learning CoT approaches \cite{wei2022chain, wang2022self}. 
Additional implementation details as well as the prompt design and sample outputs are provided in Appendix \ref{ap:implement} and \ref{ap:prompt} respectively.

\vspace*{-0.1cm}
\subsection{Baseline}
\vspace*{-0.1cm}
\paragraph{Chain-of-Thought (CoT) Approach Comparison} We compare our proposed method with the current relevant state-of-the-art CoT approaches:

\noindent $\bullet$ \textbf{Direct Prompt}: Naive prompting to generate the Logic Form given the intent and slot vocabulary. \\
\noindent $\bullet$ \textbf{CoT} \cite{wei2022chain}: Automatic generation of series of intermediate reasoning steps from LLMs  \\
\noindent $\bullet$ \textbf{SC-CoT} \cite{wang2022self}: Enhanced CoT via majority voting among multiple reasoning paths. \\
\noindent $\bullet$ \textbf{Complex-CoT} \cite{fu2022complexity} : Enhanced CoT by selecting and measuring the consistency of the most complex samples. In our case, we leverage the longest output as the complexity measure.  \\
\noindent $\bullet$ \textbf{Least-to-Most} \cite{zhou2022least} : Enhanced CoT by first automatically decomposing the in-hand problems into series of simpler sub-problems, and then solving each sub-problem sequentially.  

\noindent $\bullet$ \textbf{Plan-and-Solve} \cite{wang-etal-2023-plan} : Enhanced CoT by guiding LLMs to devise the plan before solving the problems by prompting \textit{``Let’s first understand the problem and devise a plan to solve the problem. Then, let’s carry out the plan and solve the problem step by step.''}  

 \vspace*{-0.1cm}
\paragraph{Fine-tuning (FT) Approach Comparison} As one of the early studies in leveraging LLM for NLU tasks, we also conduct additional comparisons with traditional FT approaches. Specifically, we leverage RoBERTa PLM \cite{zhuang2021robustly} with joint Slot Filling and Intent Detection objectives \cite{li-etal-2021-mtop} as the FT model. Unlike LLM, traditional FT operates under closed-world assumption which requires sufficient data to learn domain-specific and domain-agnostic feature extraction in multi-domain settings. For a fair comparison with LLM, we impose an essential constraint that there exist no overlapping domains between train and test domains under ZSL and FSL setting for both FT and CoT approaches. This leads to 3 different scenarios for FT approaches, including:

\noindent $\bullet$ \textbf{Fully Supervised}: Samples sharing similar domains with test sets are used for training. \\
\noindent $\bullet$ \textbf{ZSL}: We utilize samples from domains different from test domains for training. \\
\noindent $\bullet$ \textbf{FSL}: We leverage samples from domains different from test domains in conjunction with a fixed number of k-shot test domain samples.

\begin{table}[bt]
\centering
\caption{Comparison between FT and LLM approaches on MTOP dataset. }
\vspace*{-0.3cm}
\resizebox{\columnwidth}{!}{%
\begin{tabular}{|c|c||c|c|c|c|c|c||}
\hline 
Method & Assumption & Intent Acc & Slot F1 &  Frame Acc & Exact Match \\
 \hline

 RoBERTa FT & Supervised & 67.19 $\pm$ 2.90 & 75.17 $\pm$ 1.08 & 43.57$\pm$ 4.18 & 36.10 $\pm$ 1.08\\
 RoBERTa FT & ZSL & 0 & 12.68 $\pm$ 1.25 & 0 & 0 \\
 RoBERTa FT & FSL & 0 & 13.75 $\pm$ 1.22 & 0 & 0 \\
\hline

\textbf{CoF-CoT} & \textbf{ZSL}  & \textbf{57.67 $\pm$ 2.75} & \textbf{23.47 $\pm$ 4.09} & \textbf{14.33 $\pm$ 1.52} & \textbf{9.00 $\pm$ 1.00}  \\
\textbf{CoF-CoT} & \textbf{FSL}  & \textbf{61.50 $\pm$ 4.93} & \textbf{30.12 $\pm$ 3.93} & \textbf{15.00 $\pm$ 1.32} & \textbf{11.00 $\pm$ 1.61}  \\

\hline
\end{tabular}%
}
\vspace*{-0.3cm}
\label{tab:compare_ft}
\end{table}

\begin{table}[bt]
\centering
\caption{Ablation study on the effectiveness of different structured representations on MTOP dataset under zero-shot settings. \textit{CP, DP, AMR} denote \textit{Constituency Parsing, Dependency Parsing and Abstract Meaning Representation} respectively.}
\vspace*{-0.4cm}
\resizebox{\columnwidth}{!}{%
\begin{tabular}{|c||c|c|c|c|c|c||}
\hline 
 & Intent Acc & Slot F1 &  Frame Acc & Exact Match \\
 \hline
CoT (w/o structure) & 57.16 $\pm$ 3.69 & 17.50 $\pm$  2.92  & 12.16 $\pm$ 1.61  & 4.67 $\pm$ 3.33  \\
\hline

CP-CoT & 57.33 $\pm$ 3.25 & 19.34 $\pm$ 3.34 &   13.16 $\pm$ 1.04 & 5.50 $\pm$ 1.32 \\

DP-CoT & 57.50 $\pm$ 3.01 & 17.83 $\pm$ 2.53 & 12.67 $\pm$ 1.04 & 5.83 $\pm$ 2.08 \\
\hline
\textbf{AMR-CoT} & \textbf{57.67 $\pm$ 2.75} & \textbf{23.47 $\pm$ 4.09} & \textbf{14.33 $\pm$ 1.52} & \textbf{9.00 $\pm$ 1.00} \\
\hline
\end{tabular}%
}
\vspace*{-0.4cm}
\label{tab:ablation_structure}
\end{table}

\begin{table*}[bt]
\centering
\caption{Ablation study of step ordering
on MASSIVE dataset. CoF and FoC denote Coarse-to-Fine-grained and Fine-to-Coarse-grained order respectively.}
\vspace*{-0.3cm}
\resizebox{\textwidth}{!}{%
\begin{tabular}{|l|c||c|c|c|c|c|c||}
\hline 
Method & Assumption & Intent Acc & Slot F1 &  Frame Acc & Exact Match \\

 \hline

Random-CoT & Random Order 
 & 80.67 $\pm$3.60 & 27.14 $\pm$ 2.47 & 26.50 $\pm$1.80 & 16.50 $\pm$ 1.04 

\\
 
FoC-CoT & FoC order
 & 83.00 $\pm$ 2.88 &32.11 $\pm$ 2.50 & 28.50 $\pm$ 3.21 & 18.00 $\pm$ 3.50

\\

\hline


CoF-CoT (w/o step 1) & No AMR & 81.50 $\pm$ 4.36 & 33.68 $\pm$ 2.40 & 27.50 $\pm$ 2.65 & 18.00 $\pm$ 0.76 \\
CoF-CoT (w/o step 2) & No intent & 78.17 $\pm$ 4.80 & 27.66 $\pm$ 1.93 & 23.50 $\pm$ 2.78 & 14.50 $\pm$ 2.25 \\
CoF-CoT (w/o step 3) & No separate KP & 82.33 $\pm$ 1.04 & 34.63 $\pm$ 3.10 & 32.83 $\pm$ 2.47 & 23.00 $\pm$ 1.80 \\
CoF-CoT (w/o step 4) & No separate slot prediction for KP & 79.17 $\pm$ 4.01 & 32.92 $\pm$ 5.02 & 31.50 $\pm$ 3.50 & 21.83 $\pm$ 3.17 \\
CoF-CoT (w/o step 3+4) & No separate slot prediction & 81.33 $\pm$ 4.19 & 31.31 $\pm$3.77 & 27.67 $\pm$ 5.34 & 21.00 $\pm$ 4.92 \\
\hline 

\textbf{CoF-CoT} & \textbf{CoF order (Full)}  & \textbf{89.00 $\pm$ 2.29} & \textbf{38.66 $\pm$ 3.25} & \textbf{33.17 $\pm$ 4.04} & \textbf{25.50 $\pm$ 2.64} \\

\hspace{0.2cm} - Conditioning & No domain 
& 84.50 $\pm$ 2.75 & 36.80 $\pm$ 2.08  & 32.50 $\pm$ 1.73 & 24.83 $\pm$ 0.58 \\
\hline
\end{tabular}%
}
 \vspace*{-0.5cm}
\label{tab:ablation_condition}
\end{table*}

\vspace*{-0.2cm}
\section{Result \& Discussion}
\label{sec:result}
\vspace*{-0.2cm}
As observed in Table \ref{tab:main_result}, our proposed CoF-CoT achieves state-of-the-art performance across different evaluation metrics on MASSIVE and MTOP datasets under both zero-shot and few-shot settings. The performance gain over the most competitive baseline is more significant in terms of Intent Accuracy (25\% and 15.34\% improvements on MTOP and MASSIVE respectively in zero-shot settings). Additional case studies presented in Appendix  \ref{ap:qualitative} further demonstrate the effectiveness of CoF-CoT.    

In addition, we observe consistent improvements of different CoT variants over the Direct Prompt. It implies that CoT prompting allows the model to reason over multiple steps and learn the connections between different NLU tasks more effectively.

In comparison with MASSIVE, performance of all methods is significantly lower on MTOP. It is mainly due to the more complex Logic Form structures existent in MTOP. It is noticeable that MASSIVE datasets contain samples of fewer average number of slots, leading to significantly better performance on Semantic Parsing tasks (i.e. Frame Accuracy and Exact Match). 


Our CoF-CoT shares certain degrees of similarities with Least-to-Most \cite{zhou2022least}, Plan-and-Solve prompting \cite{wang-etal-2023-plan}. However, unlike the two aforementioned baselines that rely heavily on the existent pre-trained knowledge of LLMs, CoF-CoT provides a controllable number of sequential steps and conditioning inputs for each step, allowing for flexible adaptations and customizations to future downstream tasks that LLMs might not be familiar with.
\vspace*{-0.2cm}
\paragraph{Comparison with FT} 
 Under ZSL and FSL settings, the FT model suffers from the aforementioned domain gap issues. Specifically, as observed in Table \ref{tab:compare_ft}, since there exist minimal overlapping intent labels between train and test domains, without sufficient data in ZSL and FSL settings, the FT approaches are unable to learn transferable multi-domain features, leading to 0 performance in Intent Accuracy. This behavior also results in 0 performance for both Frame Acc and Exact Match as the correct intents are the prerequisites for correct semantic frame and exact match metrics. On the other hand, \textit{Fully supervised FT} approach acquires domain-specific knowledge of target domains from training data and performs the best across different evaluation metrics. However, this assumption does not directly match ZSL/FSL settings in which LLMs are currently evaluated. 
\vspace*{-0.1cm}
\paragraph{Impact of Structured Representation} Besides AMR Graph, there exist other structured representations that directly link to semantic and syntactic understanding of utterances, including DP,CP. Our empirical study presented in Table \ref{tab:ablation_structure} reveals that AMR-CoT unanimously achieves the best performance, demonstrating its effectiveness in capturing the diversity of input utterances when compared with other structured representations. 

\vspace*{-0.1cm}
\paragraph{Impact of Step Order} 
To further understand the importance of the designed CoF step order, we conduct additional ablation studies on 3 different scenarios: (1) random ordering (step 3$\rightarrow$1$\rightarrow$2$\rightarrow$4$\rightarrow$5), (2) Fine-to-Coarse (FoC) ordering (step 1$\rightarrow$3$\rightarrow$4$\rightarrow$2$\rightarrow$5), and (3) CoF ordering with hypothetical individual step removal. Table \ref{tab:ablation_condition} demonstrates CoF logical ordering yields the best performance with significant improvements on the challenging Exact Match metrics (9.00 and 7.50 points of improvement over random and FoC respectively). Random, FoC ordering together with CoF ordering with missing individual steps neglect the natural connections of problem-solving from high-level (coarse-grained) to low-level (fine-grained) tasks, leading to worse performance across different metrics. For CoF-CoT, when step 1 or step 2 is removed (no AMR or intent information), we observe the most significant performance decrease, implying the essence of coarse-grained knowledge for LLMs to solve the later sequential steps. 
\vspace*{-0.1cm}
\paragraph{Impact of Conditioning} The major advantage of our multi-step reasoning is the ability to explicitly condition the prior predictions in later steps. As observed in Table \ref{tab:ablation_condition}, conditioning prior knowledge in multi-step reasoning improves  the overall performance of CoF-CoT across different metrics with the most significant gain in Intent Accuracy (+4.50\%). This observation implies the importance of conditioning the appropriate information on CoT for an improved performance of LLMs under challenging zero-shot multi-domain settings.  

 \vspace*{-0.1cm}
\section{Conclusion}
\label{sec:conclusion}
\vspace*{-0.2cm}
In this work, we conduct a preliminary study of LLMs' capability in multi-grained NLU tasks of Dialogue systems. Moreover, motivated by CoT, we propose a novel CoF-CoT approach aiming to break down NLU tasks into multiple reasoning steps where (1) LLMs can learn to acquire and leverage concepts from different granularities of NLU tasks, (2) additional AMR structured representation can be integrated and leveraged throughout the multi-step reasoning. We empirically demonstrate the effectiveness of CoF-CoT in improving LLMs capability in multi-grained NLU tasks under both zero-shot and few-shot multi-domain settings. 



\section*{Limitations}
Our empirical study is restricted to English NLU data. It is partially due to the existent English-bias of Abstract Meaning Representation (AMR) structure \cite{banarescu2013abstract}. We leave the adaptation of the CoF-CoT to multilingual settings \cite{nguyen2019cross, qin2022gl,nguyen-etal-2023-enhancing} as future directions for our work.

Our work is empirically studied on the Flat Logic Form representation. In other words, Logic Form only includes one intent followed by a set of slot sequences. There are two major rationales for our empirical scope. Firstly, as the early preliminary study on multi-grained NLU tasks which unify both Semantic Parsing and NLU perspectives, we design a small and controllable scope for the experiments. Secondly, as most NLU datasets including MASSIVE \cite{fitzgerald2022massive} are restricted to single-intent utterances, Flat Logic Form is a viable candidate reconciliating between traditional NLU and Semantic Parsing evaluations. We leave explorations on the more challenging Nested Logic Form where utterances might contain multiple intents for future work.  

\section{Acknowledgement}
We thank the anonymous reviewers for their constructive feedback which we incorporated in the final version of this manuscript.

This work is supported in part by NSF under grant III-2106758.

\bibliography{custom}

\begin{thebibliography}{35}
\expandafter\ifx\csname natexlab\endcsname\relax\def\natexlab#1{#1}\fi

\bibitem[{Bai et~al.(2021)Bai, Wang, Chen, Yang, Bai, Yu, and Tong}]{bai-etal-2021-syntax}
Jiangang Bai, Yujing Wang, Yiren Chen, Yaming Yang, Jing Bai, Jing Yu, and Yunhai Tong. 2021.
\newblock \href {https://doi.org/10.18653/v1/2021.eacl-main.262} {Syntax-{BERT}: Improving pre-trained transformers with syntax trees}.
\newblock In \emph{Proceedings of the 16th Conference of the European Chapter of the Association for Computational Linguistics: Main Volume}, pages 3011--3020, Online. Association for Computational Linguistics.

\bibitem[{Banarescu et~al.(2013)Banarescu, Bonial, Cai, Georgescu, Griffitt, Hermjakob, Knight, Koehn, Palmer, and Schneider}]{banarescu2013abstract}
Laura Banarescu, Claire Bonial, Shu Cai, Madalina Georgescu, Kira Griffitt, Ulf Hermjakob, Kevin Knight, Philipp Koehn, Martha Palmer, and Nathan Schneider. 2013.
\newblock Abstract meaning representation for sembanking.
\newblock In \emph{Proceedings of the 7th linguistic annotation workshop and interoperability with discourse}, pages 178--186.

\bibitem[{Bastianelli et~al.(2020)Bastianelli, Vanzo, Swietojanski, and Rieser}]{bastianelli-etal-2020-slurp}
Emanuele Bastianelli, Andrea Vanzo, Pawel Swietojanski, and Verena Rieser. 2020.
\newblock \href {https://doi.org/10.18653/v1/2020.emnlp-main.588} {{SLURP}: A spoken language understanding resource package}.
\newblock In \emph{Proceedings of the 2020 Conference on Empirical Methods in Natural Language Processing (EMNLP)}, pages 7252--7262, Online. Association for Computational Linguistics.

\bibitem[{Bugliarello and Okazaki(2020)}]{bugliarello-okazaki-2020-enhancing}
Emanuele Bugliarello and Naoaki Okazaki. 2020.
\newblock \href {https://doi.org/10.18653/v1/2020.acl-main.147} {Enhancing machine translation with dependency-aware self-attention}.
\newblock In \emph{Proceedings of the 58th Annual Meeting of the Association for Computational Linguistics}, pages 1618--1627, Online. Association for Computational Linguistics.

\bibitem[{Casanueva et~al.(2022)Casanueva, Vuli{\'c}, Spithourakis, and Budzianowski}]{casanueva2022nlu++}
I{\~n}igo Casanueva, Ivan Vuli{\'c}, Georgios Spithourakis, and Pawe{\l} Budzianowski. 2022.
\newblock Nlu++: A multi-label, slot-rich, generalisable dataset for natural language understanding in task-oriented dialogue.
\newblock In \emph{Findings of the Association for Computational Linguistics: NAACL 2022}, pages 1998--2013.

\bibitem[{Chowdhery et~al.(2022)Chowdhery, Narang, Devlin, Bosma, Mishra, Roberts, Barham, Chung, Sutton, Gehrmann et~al.}]{chowdhery2022palm}
Aakanksha Chowdhery, Sharan Narang, Jacob Devlin, Maarten Bosma, Gaurav Mishra, Adam Roberts, Paul Barham, Hyung~Won Chung, Charles Sutton, Sebastian Gehrmann, et~al. 2022.
\newblock Palm: Scaling language modeling with pathways.
\newblock \emph{arXiv preprint arXiv:2204.02311}.

\bibitem[{Firdaus et~al.(2019)Firdaus, Kumar, Ekbal, and Bhattacharyya}]{firdaus2019multi}
Mauajama Firdaus, Ankit Kumar, Asif Ekbal, and Pushpak Bhattacharyya. 2019.
\newblock A multi-task hierarchical approach for intent detection and slot filling.
\newblock \emph{Knowledge-Based Systems}, 183:104846.

\bibitem[{FitzGerald et~al.(2022)FitzGerald, Hench, Peris, Mackie, Rottmann, Sanchez, Nash, Urbach, Kakarala, Singh, Ranganath, Crist, Britan, Leeuwis, Tur, and Natarajan}]{fitzgerald2022massive}
Jack FitzGerald, Christopher Hench, Charith Peris, Scott Mackie, Kay Rottmann, Ana Sanchez, Aaron Nash, Liam Urbach, Vishesh Kakarala, Richa Singh, Swetha Ranganath, Laurie Crist, Misha Britan, Wouter Leeuwis, Gokhan Tur, and Prem Natarajan. 2022.
\newblock \href {http://arxiv.org/abs/2204.08582} {Massive: A 1m-example multilingual natural language understanding dataset with 51 typologically-diverse languages}.

\bibitem[{Fu et~al.(2022)Fu, Peng, Sabharwal, Clark, and Khot}]{fu2022complexity}
Yao Fu, Hao Peng, Ashish Sabharwal, Peter Clark, and Tushar Khot. 2022.
\newblock Complexity-based prompting for multi-step reasoning.
\newblock \emph{arXiv preprint arXiv:2210.00720}.

\bibitem[{Kamath and Das()}]{kamathsurvey}
Aishwarya Kamath and Rajarshi Das.
\newblock A survey on semantic parsing.
\newblock In \emph{Automated Knowledge Base Construction (AKBC)}.

\bibitem[{Khattab et~al.(2022)Khattab, Santhanam, Li, Hall, Liang, Potts, and Zaharia}]{khattab2022demonstrate}
Omar Khattab, Keshav Santhanam, Xiang~Lisa Li, David Hall, Percy Liang, Christopher Potts, and Matei Zaharia. 2022.
\newblock Demonstrate-search-predict: Composing retrieval and language models for knowledge-intensive nlp.
\newblock \emph{arXiv preprint arXiv:2212.14024}.

\bibitem[{Li et~al.(2021)Li, Arora, Chen, Gupta, Gupta, and Mehdad}]{li-etal-2021-mtop}
Haoran Li, Abhinav Arora, Shuohui Chen, Anchit Gupta, Sonal Gupta, and Yashar Mehdad. 2021.
\newblock \href {https://doi.org/10.18653/v1/2021.eacl-main.257} {{MTOP}: A comprehensive multilingual task-oriented semantic parsing benchmark}.
\newblock In \emph{Proceedings of the 16th Conference of the European Chapter of the Association for Computational Linguistics: Main Volume}, pages 2950--2962, Online. Association for Computational Linguistics.

\bibitem[{Nguyen and Rohrbaugh(2019)}]{nguyen2019cross}
Hoang Nguyen and Gene Rohrbaugh. 2019.
\newblock Cross-lingual genre classification using linguistic groupings.
\newblock \emph{Journal of Computing Sciences in Colleges}, 34(3):91--96.

\bibitem[{Nguyen et~al.(2023{\natexlab{a}})Nguyen, Zhang, Liu, and Yu}]{nguyen-etal-2023-slot}
Hoang Nguyen, Chenwei Zhang, Ye~Liu, and Philip Yu. 2023{\natexlab{a}}.
\newblock \href {https://aclanthology.org/2023.sigdial-1.44} {Slot induction via pre-trained language model probing and multi-level contrastive learning}.
\newblock In \emph{Proceedings of the 24th Meeting of the Special Interest Group on Discourse and Dialogue}, pages 470--481, Prague, Czechia. Association for Computational Linguistics.

\bibitem[{Nguyen et~al.(2020)Nguyen, Zhang, Xia, and Philip}]{nguyen2020dynamic}
Hoang Nguyen, Chenwei Zhang, Congying Xia, and S~Yu Philip. 2020.
\newblock Dynamic semantic matching and aggregation network for few-shot intent detection.
\newblock In \emph{Findings of the Association for Computational Linguistics: EMNLP 2020}, pages 1209--1218.

\bibitem[{Nguyen et~al.(2023{\natexlab{b}})Nguyen, Zhang, Zhang, Rohrbaugh, and Yu}]{nguyen-etal-2023-enhancing}
Hoang Nguyen, Chenwei Zhang, Tao Zhang, Eugene Rohrbaugh, and Philip Yu. 2023{\natexlab{b}}.
\newblock \href {https://doi.org/10.18653/v1/2023.findings-acl.583} {Enhancing cross-lingual transfer via phonemic transcription integration}.
\newblock In \emph{Findings of the Association for Computational Linguistics: ACL 2023}, pages 9163--9175, Toronto, Canada. Association for Computational Linguistics.

\bibitem[{Qin et~al.(2022)Qin, Chen, Xie, Li, Lou, Che, and Kan}]{qin2022gl}
Libo Qin, Qiguang Chen, Tianbao Xie, Qixin Li, Jian-Guang Lou, Wanxiang Che, and Min-Yen Kan. 2022.
\newblock Gl-clef: A global--local contrastive learning framework for cross-lingual spoken language understanding.
\newblock In \emph{Proceedings of the 60th Annual Meeting of the Association for Computational Linguistics (Volume 1: Long Papers)}, pages 2677--2686.

\bibitem[{Shou et~al.(2022)Shou, Jiang, and Lin}]{shou-etal-2022-amr}
Ziyi Shou, Yuxin Jiang, and Fangzhen Lin. 2022.
\newblock \href {https://doi.org/10.18653/v1/2022.findings-acl.244} {{AMR-DA}: {D}ata augmentation by {A}bstract {M}eaning {R}epresentation}.
\newblock In \emph{Findings of the Association for Computational Linguistics: ACL 2022}, pages 3082--3098, Dublin, Ireland. Association for Computational Linguistics.

\bibitem[{Wang et~al.(2023)Wang, Xu, Lan, Hu, Lan, Lee, and Lim}]{wang-etal-2023-plan}
Lei Wang, Wanyu Xu, Yihuai Lan, Zhiqiang Hu, Yunshi Lan, Roy Ka-Wei Lee, and Ee-Peng Lim. 2023.
\newblock \href {https://doi.org/10.18653/v1/2023.acl-long.147} {Plan-and-solve prompting: Improving zero-shot chain-of-thought reasoning by large language models}.
\newblock In \emph{Proceedings of the 61st Annual Meeting of the Association for Computational Linguistics (Volume 1: Long Papers)}, pages 2609--2634, Toronto, Canada. Association for Computational Linguistics.

\bibitem[{Wang et~al.(2022)Wang, Wei, Schuurmans, Le, Chi, Narang, Chowdhery, and Zhou}]{wang2022self}
Xuezhi Wang, Jason Wei, Dale Schuurmans, Quoc Le, Ed~Chi, Sharan Narang, Aakanksha Chowdhery, and Denny Zhou. 2022.
\newblock Self-consistency improves chain of thought reasoning in language models.
\newblock \emph{arXiv preprint arXiv:2203.11171}.

\bibitem[{Wei et~al.(2022)Wei, Wang, Schuurmans, Bosma, Chi, Le, and Zhou}]{wei2022chain}
Jason Wei, Xuezhi Wang, Dale Schuurmans, Maarten Bosma, Ed~Chi, Quoc Le, and Denny Zhou. 2022.
\newblock Chain of thought prompting elicits reasoning in large language models.
\newblock \emph{arXiv preprint arXiv:2201.11903}.

\bibitem[{Weld et~al.(2022)Weld, Huang, Long, Poon, and Han}]{weld2022survey}
Henry Weld, Xiaoqi Huang, Siqu Long, Josiah Poon, and Soyeon~Caren Han. 2022.
\newblock A survey of joint intent detection and slot filling models in natural language understanding.
\newblock \emph{ACM Computing Surveys}, 55(8):1--38.

\bibitem[{Wu et~al.(2020)Wu, Ding, Lu, and Xie}]{wu-etal-2020-slotrefine}
Di~Wu, Liang Ding, Fan Lu, and Jian Xie. 2020.
\newblock \href {https://doi.org/10.18653/v1/2020.emnlp-main.152} {{S}lot{R}efine: A fast non-autoregressive model for joint intent detection and slot filling}.
\newblock In \emph{Proceedings of the 2020 Conference on Empirical Methods in Natural Language Processing (EMNLP)}, pages 1932--1937, Online. Association for Computational Linguistics.

\bibitem[{Xia et~al.(2020)Xia, Zhang, Nguyen, Zhang, and Yu}]{xia2020cg}
Congying Xia, Chenwei Zhang, Hoang Nguyen, Jiawei Zhang, and Philip Yu. 2020.
\newblock Cg-bert: Conditional text generation with bert for generalized few-shot intent detection.
\newblock \emph{arXiv preprint arXiv:2004.01881}.

\bibitem[{Xie et~al.(2022)Xie, Wu, Shi, Zhong, Scholak, Yasunaga, Wu, Zhong, Yin, Wang, Zhong, Wang, Li, Boyle, Ni, Yao, Radev, Xiong, Kong, Zhang, Smith, Zettlemoyer, and Yu}]{xie-etal-2022-unifiedskg}
Tianbao Xie, Chen~Henry Wu, Peng Shi, Ruiqi Zhong, Torsten Scholak, Michihiro Yasunaga, Chien-Sheng Wu, Ming Zhong, Pengcheng Yin, Sida~I. Wang, Victor Zhong, Bailin Wang, Chengzu Li, Connor Boyle, Ansong Ni, Ziyu Yao, Dragomir Radev, Caiming Xiong, Lingpeng Kong, Rui Zhang, Noah~A. Smith, Luke Zettlemoyer, and Tao Yu. 2022.
\newblock \href {https://aclanthology.org/2022.emnlp-main.39} {{U}nified{SKG}: Unifying and multi-tasking structured knowledge grounding with text-to-text language models}.
\newblock In \emph{Proceedings of the 2022 Conference on Empirical Methods in Natural Language Processing}, pages 602--631, Abu Dhabi, United Arab Emirates. Association for Computational Linguistics.

\bibitem[{Xu et~al.(2021)Xu, Guo, Tang, Su, Shou, Gong, Zhong, Quan, Jiang, and Duan}]{xu2021syntax}
Zenan Xu, Daya Guo, Duyu Tang, Qinliang Su, Linjun Shou, Ming Gong, Wanjun Zhong, Xiaojun Quan, Daxin Jiang, and Nan Duan. 2021.
\newblock Syntax-enhanced pre-trained model.
\newblock In \emph{Proceedings of the 59th Annual Meeting of the Association for Computational Linguistics and the 11th International Joint Conference on Natural Language Processing (Volume 1: Long Papers)}, pages 5412--5422.

\bibitem[{Yao et~al.(2022)Yao, Zhao, Yu, Du, Shafran, Narasimhan, and Cao}]{yao2022react}
Shunyu Yao, Jeffrey Zhao, Dian Yu, Nan Du, Izhak Shafran, Karthik Narasimhan, and Yuan Cao. 2022.
\newblock React: Synergizing reasoning and acting in language models.
\newblock \emph{arXiv preprint arXiv:2210.03629}.

\bibitem[{Zhang et~al.(2019)Zhang, Li, Du, Fan, and Philip}]{zhang2019joint}
Chenwei Zhang, Yaliang Li, Nan Du, Wei Fan, and S~Yu Philip. 2019.
\newblock Joint slot filling and intent detection via capsule neural networks.
\newblock In \emph{Proceedings of the 57th Annual Meeting of the Association for Computational Linguistics}, pages 5259--5267.

\bibitem[{Zhang et~al.(2020)Zhang, Wu, Zhou, Duan, Zhao, and Wang}]{zhang2020sg}
Zhuosheng Zhang, Yuwei Wu, Junru Zhou, Sufeng Duan, Hai Zhao, and Rui Wang. 2020.
\newblock Sg-net: Syntax-guided machine reading comprehension.
\newblock In \emph{Proceedings of the AAAI Conference on Artificial Intelligence}, volume~34, pages 9636--9643.

\bibitem[{Zhao et~al.(2023{\natexlab{a}})Zhao, Li, Joty, Qin, and Bing}]{zhao2023verify}
Ruochen Zhao, Xingxuan Li, Shafiq Joty, Chengwei Qin, and Lidong Bing. 2023{\natexlab{a}}.
\newblock Verify-and-edit: A knowledge-enhanced chain-of-thought framework.
\newblock \emph{arXiv preprint arXiv:2305.03268}.

\bibitem[{Zhao et~al.(2023{\natexlab{b}})Zhao, Zhou, Li, Tang, Wang, Hou, Min, Zhang, Zhang, Dong et~al.}]{zhao2023survey}
Wayne~Xin Zhao, Kun Zhou, Junyi Li, Tianyi Tang, Xiaolei Wang, Yupeng Hou, Yingqian Min, Beichen Zhang, Junjie Zhang, Zican Dong, et~al. 2023{\natexlab{b}}.
\newblock A survey of large language models.
\newblock \emph{arXiv preprint arXiv:2303.18223}.

\bibitem[{Zhou et~al.(2022)Zhou, Sch{\"a}rli, Hou, Wei, Scales, Wang, Schuurmans, Cui, Bousquet, Le et~al.}]{zhou2022least}
Denny Zhou, Nathanael Sch{\"a}rli, Le~Hou, Jason Wei, Nathan Scales, Xuezhi Wang, Dale Schuurmans, Claire Cui, Olivier Bousquet, Quoc~V Le, et~al. 2022.
\newblock Least-to-most prompting enables complex reasoning in large language models.
\newblock In \emph{The Eleventh International Conference on Learning Representations}.

\bibitem[{Zhou et~al.(2023)Zhou, Iacobacci, and Minervini}]{zhou-etal-2023-xqa}
Han Zhou, Ignacio Iacobacci, and Pasquale Minervini. 2023.
\newblock \href {https://aclanthology.org/2023.findings-eacl.73} {{XQA}-{DST}: Multi-domain and multi-lingual dialogue state tracking}.
\newblock In \emph{Findings of the Association for Computational Linguistics: EACL 2023}, pages 999--1009, Dubrovnik, Croatia. Association for Computational Linguistics.

\bibitem[{Zhuang et~al.(2021)Zhuang, Wayne, Ya, and Jun}]{zhuang2021robustly}
Liu Zhuang, Lin Wayne, Shi Ya, and Zhao Jun. 2021.
\newblock A robustly optimized bert pre-training approach with post-training.
\newblock In \emph{Proceedings of the 20th Chinese National Conference on Computational Linguistics}, pages 1218--1227.

\bibitem[{Zoph et~al.(2022)Zoph, Raffel, Schuurmans, Yogatama, Zhou, Metzler, Chi, Wei, Dean, Fedus, Bosma, Vinyals, Liang, Borgeaud, Hashimoto, and Tay}]{52065}
Barret Zoph, Colin Raffel, Dale Schuurmans, Dani Yogatama, Denny Zhou, Don Metzler, Ed~H. Chi, Jason Wei, Jeff Dean, Liam~B. Fedus, Maarten~Paul Bosma, Oriol Vinyals, Percy Liang, Sebastian Borgeaud, Tatsunori~B. Hashimoto, and Yi~Tay. 2022.
\newblock Emergent abilities of large language models.
\newblock \emph{TMLR}.

\end{thebibliography}
\bibliographystyle{acl_natbib}

\newpage
\appendix

\section{Connections between Semantic Parsing and NLU Tasks via Logic Form}
\label{ap:logic_form}
Logic Form not only captures the coarse-grained intent labels and fine-grained slot labels of the utterances but also encapsulates the implicit connections between slots and intents. 

As observed in Table \ref{tab:ex_mtop}, Logic Form is constructed as the flattened representation of the dependency structure between intents and slot sequences. Semantic Frame constructed as intent type(s) followed by a sequence of slot types can be directly extracted from the Logic Form. In addition, via the Logic Form, the coarse-grained intent label CREATE\_REMINDER, fine-grained slot TODO, DATE\_TIME labels together the respective slot values (\textit{message Mike}, \textit{at 7pm tonight}) can all be extracted and converted to appropriate format (i.e. BIO format as the traditional sequence labeling ground truths \cite{zhang2019joint}). Therefore, Logic Form can be considered the unified label format to bridge the gap between Semantic Parsing \cite{li-etal-2021-mtop,xie-etal-2022-unifiedskg} and traditional Intent Detection and Slot Filling tasks in NLU systems \cite{xia2020cg,nguyen2020dynamic,casanueva2022nlu++}. 

\section{Dataset Details}
\label{ap:dataset}
\begin{table}[tb]
\centering
\caption{Details of MTOP and MASSIVE datasets}
 \vspace*{-0.2cm}
\resizebox{0.8\columnwidth}{!}{%
\begin{tabular}{|c|c|c|}
\hline 
Dataset & \textbf{MTOP} & \textbf{MASSIVE} \\
\hline 
\# Domains & 11 & 18\\
\# Train Domains & 8 & 14\\
\# Test Domains & 3 & 4\\
\# Intents & 117 & 60 \\
\# Slots & 78 & 55 \\
Sentence Length & 6.14 $\pm$ 2.30 & 6.34 $\pm$ 2.94 \\
\# Slots per sample & 1.87 $\pm$ 0.81 & 0.73 $\pm$ 0.63 \\ 
\hline
\end{tabular}%
}
 \vspace*{-0.4cm}
\label{tab:dataset}
\end{table}

We provide the details of MTOP and MASSIVE datasets in Table \ref{tab:dataset}. As compared to MASSIVE, MTOP dataset not only contains more slot types and intent types but also tends to cover more slot types per sample in the Logic Form. This challenging characteristic explains the consistent lower performance across all methods on MTOP when compared to MASSIVE as observed in Section \ref{sec:result}. 

\section{Implementation Details}
\label{ap:implement}
As the proposed step-by-step reasoning can be applied to any LLMs, our proposed method is LLM-agnostic which is empirically studied in Appendix \ref{ap:llm_agnostic}. For simplicity and consistency, in our main empirical study, we leverage \textit{gpt-3.5-turbo} from OpenAI as the base LLM model. Following \cite{wang2022self}, we set the decoding temperature T=0.7 and number of outputs n=10. 

As domain names provide essential clues for language models in multi-domain settings for multi-grained NLU tasks \cite{zhou-etal-2023-xqa},  to safeguard the fairness in baseline comparisons, we consistently include the \textit{domain name} in the input prompts for all baselines unless stated otherwise. Specifically, the only exception is presented in Table \ref{tab:ablation_condition} for \textit{CoF-CoT(CoF order)-Conditioning}. 

For few-shot (i.e. k-shot) learning settings, we randomly sample k examples
and manually prepare the necessary labels for different baseline variants. We experiment with k=5 in our empirical study. 

\paragraph{Domains of Demonstration Samples} To replicate a more realistic scenario where the domains of k-shot demonstration samples are generally unknown, we assume that k-shot demonstration samples come from different domains from the test samples. The relaxation of constraints on the assumption regarding the domain similarity between demonstration samples and test samples allows for broader applications and encourages LLMs to accumulate and extract the true semantic knowledge from k-shot demonstrations and avoid overfitting any specific domains. For completeness, we also conduct additional empirical studies to compare the FSL performance of CoF-CoT under both scenarios: (1) k demonstration samples are from the same domain as test samples, (2) k demonstrations are drawn from different domains from the test samples. As observed in Table \ref{tab:ksample}, additional constraint of similar domains between k-shot demonstration samples and test samples leads to improvements in the evaluation performance across NLU and Semantic Parsing tasks. This might be intuitive since LLMs can extract domain-relevant information from the given k domain-similar samples to assist with inference process on test samples.  

\begin{table*}[tb]
\centering
\caption{FSL Results of CoF-CoT with k-shot demonstration samples selected from different and similar domains in comparison with domains of test samples on MTOP dataset.}
 \vspace*{-0.2cm}
\resizebox{\textwidth}{!}{%
\begin{tabular}{|c|c|c|c|c|c|}
\hline 
Method & \textbf{Assumption} & Intent Acc & Slot F1 & Frame Acc & Exact Match \\
\hline 
CoF-CoT & k domain-different samples & 61.50 $\pm$ 4.93 & 30.12 $\pm$ 3.93 &  15.00 $\pm$1.32 & 11.0 $\pm$ 1.61 \\
CoF-CoT & k domain-similar samples & 70.00 $\pm$ 1.33 & 38.16 $\pm$ 5.42 & 20.50 $\pm$ 2.00 & 15.00 $\pm$ 1.00 \\

\hline
\end{tabular}%
}
\label{tab:ksample}
\end{table*}

\section{Prompt Design}
\label{ap:prompt}
\begin{figure*}[t]

    \centering
   \includegraphics[trim={0.4cm 0.0cm 2.0cm 0.0cm},clip, width=\textwidth]{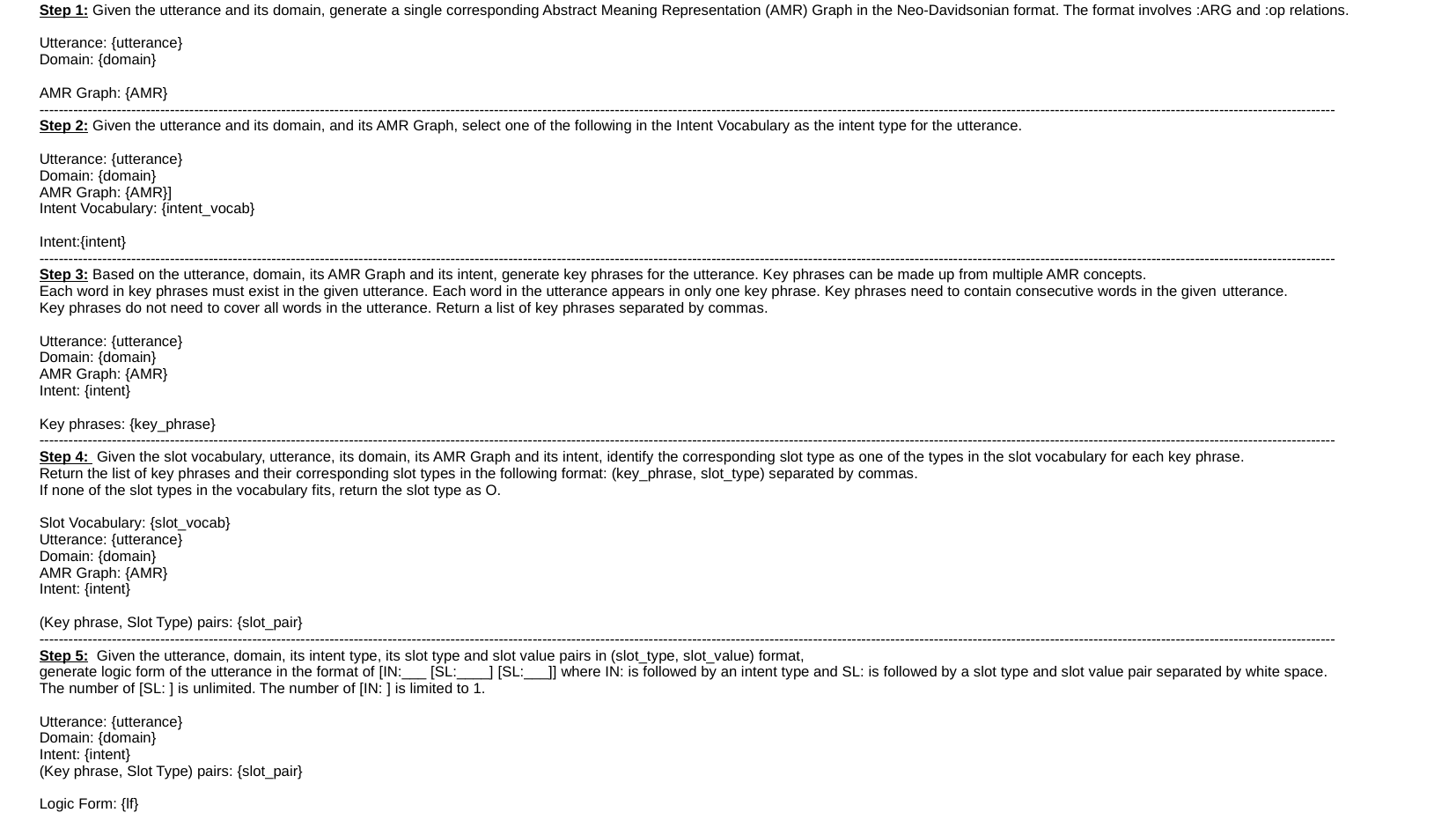}
    \caption{CoF-CoT Prompt Design Template. \{.\} denotes the placeholder argument. 
    }
    \label{fig:prompt_template}
    \vspace*{-0.5cm}
\end{figure*}

\begin{figure*}[t]
    \centering
   \includegraphics[trim={2.3cm 3.8cm 7.5cm 2.0cm},clip, width=\textwidth]{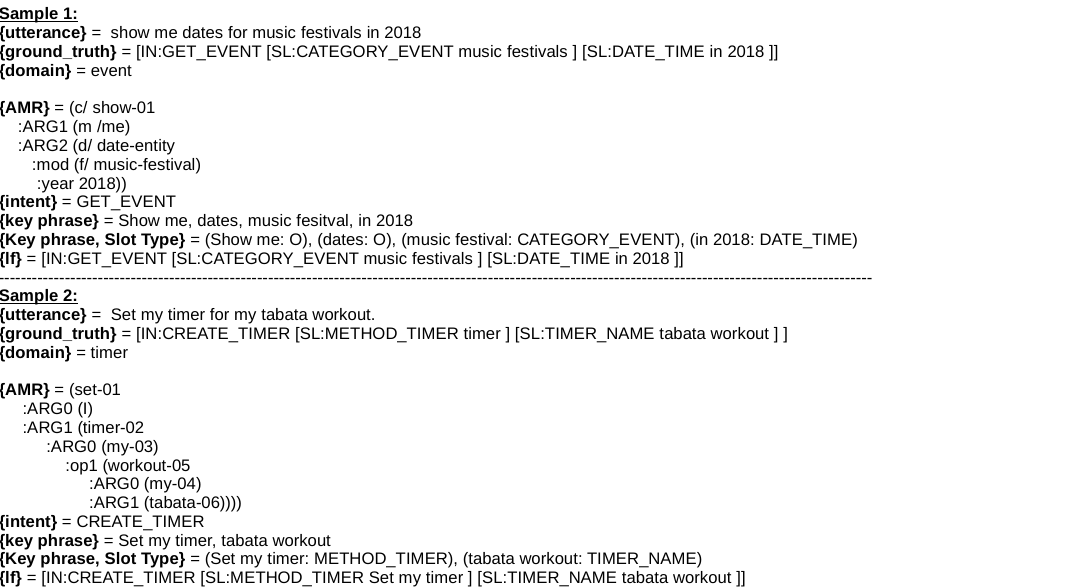}
    \caption{Sample output from our CoF-CoT. \{.\} denotes the placeholder corresponding to template in Figure \ref{fig:prompt_template}. 
    }
    \label{fig:output_sample}
    \vspace*{-0.5cm}
\end{figure*}

Prompts for individual steps of our CoF-CoT are presented in Figure \ref{fig:prompt_template}. Additional output samples are also provided in Figure \ref{fig:output_sample}. 

\section{Qualitative Case Study}
\label{ap:qualitative}
\begin{figure*}[t]
    \centering
   \includegraphics[trim={0.4cm 8.0cm 10.5cm 0.0cm},clip, width=\textwidth]{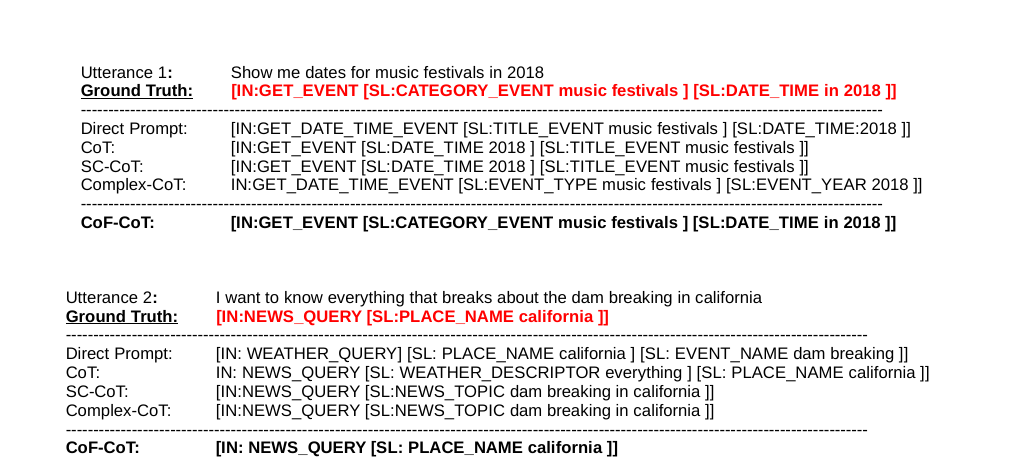}
    \caption{Qualitative Case Study among baseline variants and the proposed CoF-CoT. \textcolor{red}{Ground Truth} is shown in \textcolor{red}{red}.
    }
    \label{fig:qualitative}
    \vspace*{-0.5cm}
\end{figure*}

We present additional Qualitative Case Study  comparing the outputs between different baseline methods and our proposed CoF-CoT in Figure \ref{fig:qualitative}.

As observed in Figure \ref{fig:qualitative}, our CoF-CoT provides the predictions closest to the ground truth while other baselines struggle to (1) generate the correct intent type (i.e. GET\_DATE\_TIME\_EVENT intent type from Direct Prompt in comparison with GET\_EVENT intent from ground truth) (2) identify the correct slot values (i.e. \textit{everything}  slot value generated from CoT), (3) generate the correct slot type for the corresponding slot values. (i.e. EVENT\_TYPE slot type for \textit{music festivals} slot values from Complex-CoT instead of CATEGORY\_EVENT slot type). 
\begin{table*}[t]
\centering
\caption{Sample utterance with its Logic Form under both Semantic Parsing and NLU tasks' metrics. // denotes the separation between  tokens of the given utterance. }
 \vspace*{-0.2cm}
\resizebox{\textwidth}{!}{%
\begin{tabular}{|c|c|c|c|c|}
\hline 
& \textbf{Metric} & \textbf{Granularity Level} &\textbf{Format} & \textbf{Ground Truth} \\
\hline 
Input Sentence & -- & -- & -- & Set // up// a // reminder // to // \textcolor{red}{message} // \textcolor{red}{mike} // \textcolor{blue}{at} // \textcolor{blue}{7pm} // \textcolor{blue}{tonight} \\
Logic Form & -- & -- & -- & [\textcolor{orange}{IN:CREATE\_REMINDER} [\textcolor{red}{SL:TODO: message mike}] [\textcolor{blue}{SL:DATE\_TIME: at 7pm tonight}]] \\
\hline
\multirow{2}{*}{NLU Tasks}& Intent Accuracy & Coarse-grained & Intent Label & \textcolor{orange}{IN:CREATE\_REMINDER} \\

& Slot F1 & Fine-grained & BIO Slot Sequence & O // O // O // O // O // \textcolor{red}{B-TODO} // \textcolor{red}{I-TODO} // \textcolor{blue}{B-DATE\_TIME} // \textcolor{blue}{I-DATE\_TIME} // \textcolor{blue}{I-DATE\_TIME} \\

\hline 
\multirow{2}{*}{Semantic Parsing Tasks}& Frame Accuracy & Both & Logic Form & \textcolor{orange}{IN:CREATE\_REMINDER}-\textcolor{red}{SL:TODO}-\textcolor{blue}{SL:DATE\_TIME} \\

& Exact Match & Both & Logic Form & [\textcolor{orange}{IN:CREATE\_REMINDER} [\textcolor{red}{SL:TODO: message mike}] [\textcolor{blue}{SL:DATE\_TIME: at 7pm tonight}]]  \\

\hline
\end{tabular}%
}
 \vspace*{-0.4cm}
\label{tab:ex_mtop}
\end{table*}

\section{LLM-Agnostic Capability}
\label{ap:llm_agnostic}
Our proposed CoF-CoT is LLM-agnostic since the focus of the work is on the prompt design, which can be applied to any LLMs. As most LLMs rely on the high quality of the designed prompts, our proposed CoF-CoT prompt design can be used as input to any LLMs for zero-shot and in-context learning settings. This is also similarly observed in CoT \cite{wei2022chain}, SC-CoT \cite{wang2022self} and other comparable CoT methods. 
For further clarification, we report additional empirical results of our proposed CoF-CoT applied to both of the backbone PaLM \cite{chowdhery2022palm} and GPT3.5 LLMs on the MTOP dataset under both ZSL and FSL settings in Table \ref{tab:llm_agnostic}. As observed in Table \ref{tab:llm_agnostic}, CoF-CoT prompting consistently outperforms the two backbone LLMs across all NLU and Semantic Parsing tasks, demonstrating both the effectiveness and LLM-agnostic capability of our proposed CoF-CoT. 
 \begin{table*}[htb]
\centering
\caption{Experimental results on MTOP dataset under zero-shot few-shot multi-domain settings with different LLM backbone architectures (PaLM \cite{chowdhery2022palm} and GPT3.5).}
\vspace*{-0.3cm}
\resizebox{\textwidth}{!}{%
\begin{tabular}{||c||c|c||c|c||c|c||c|c||}
\hline 
\multicolumn{9}{||c||}{\textbf{MTOP}} \\
\hline 
 \textbf{Model} & \multicolumn{4}{c||}{\textbf{Zero-shot}} & \multicolumn{4}{c||}{\textbf{Few-shot}}\\
\specialrule{.1em}{0.05em}{.05em}
 & \multicolumn{2}{c||}{NLU} & \multicolumn{2}{c||}{Semantic Parsing} 
  & \multicolumn{2}{c||}{NLU} & \multicolumn{2}{c||}{Semantic Parsing} \\
  \hline 
  &  Intent Acc & Slot F1 & Frame Acc & Exact Match & Intent Acc &  Slot F1 & Frame Acc & Exact Match  \\


\specialrule{.1em}{0.05em}{.05em}

PaLM 
& 16.67 $\pm$ 2.52& 7.24 $\pm$ 1.00 & 3.17 $\pm$ 0.76 & 1.17 $\pm$ 0.76 
& 48.83 $\pm$ 4.54 & 14.24 $\pm$ 1.58 & 4.17 $\pm$ 2.02 & 2.33 $\pm$ 0.76
\\

\textbf{PaLM + CoF-CoT}
& \textbf{42.33 $\pm$ 3.33} & \textbf{13.73$\pm$ 2.88} & \textbf{4.01 $\pm$ 0.15} & \textbf{3.50 $\pm$ 1.32} 

 & \textbf{57.17 $\pm$ 3.79} & \textbf{21.47 $\pm$ 3.79} & \textbf{10.33 $\pm$ 3.06} & \textbf{6.67 $\pm$ 2.75} 
\\

\hline 
GPT3.5 
 & 31.50 $\pm$ 1.80 & 21.84 $\pm$ 2.83 & 8.33 $\pm$ 1.44 & 6.00 $\pm$ 1.32

 & 51.33 $\pm$ 3.40 & 28.35 $\pm$ 3.24 & 11.00 $\pm$ 1.80 & 8.33 $\pm$ 1.00

\\

\textbf{GPT3.5 + CoF-CoT}
& \textbf{57.67 $\pm$ 2.75} & \textbf{23.47 $\pm$ 4.09} & \textbf{14.33 $\pm$ 1.52} & \textbf{9.00 $\pm$ 1.00} 

 & \textbf{61.50 $\pm$ 4.93} & \textbf{30.12 $\pm$ 3.93} & \textbf{15.00 $\pm$ 1.32} & \textbf{11.00 $\pm$ 1.61} 
\\

\hline
\end{tabular}%
}

\label{tab:llm_agnostic}
\end{table*}

\

\end{document}